\documentclass[a4paper]{article} 

\usepackage[colorlinks,citecolor=blue,urlcolor=blue,linkcolor=blue,linktocpage=true]{hyperref}
\usepackage{url}
\usepackage{fullpage}
\usepackage{algorithm}
\usepackage{algorithmic}
\usepackage{amsmath, amsfonts, amssymb, amsthm, amsbsy, amscd, bm, bbm,mathrsfs}    
\usepackage{graphicx}
\usepackage{fancyvrb}
\usepackage{xcolor}
\usepackage{cancel}

\graphicspath{{figs/}}
\usepackage{subcaption}
\usepackage{cleveref}
\usepackage{enumitem}
\usepackage{tikz}

\setlist{leftmargin=5mm}
\usepackage{titlesec}
\usepackage{wrapfig}
\usepackage{multirow}

\newcommand{\x}{\bm{x}}
\newcommand{\p}{\bm{p}}
\renewcommand{\a}{\bm{a}}
\newcommand{\s}{\bm{s}}

\newcommand{\y}{\bm{y}}

\renewcommand{\b}{\mathbf{b}}

\newcommand{\R}{\mathbb{R}}

\newcommand{\W}{\mathbf{W}}
\renewcommand{\s}{\bm{s}}

\newtheorem{othertheorem}{othertheorem}[section]

\newtheorem{fact}[othertheorem]{Fact}

\theoremstyle{definition}

\theoremstyle{remark}
\newtheorem{remark}[othertheorem]{Remark}

\usepackage{adjustbox}

\title{Accelerating Adversarial Perturbation by 50\% with Semi-backward Propagation}
\date{}

\begin{document}

\maketitle

\begin{center}
\textbf{Zhiqi Bu}
\\
\texttt{woodyx218@gmail.com}
\end{center}

\begin{abstract}
Adversarial perturbation plays a significant role in the field of adversarial robustness, which solves a maximization problem over the input data. We show that the backward propagation of such optimization can accelerate $2\times$ (and thus the overall optimization including the forward propagation can accelerate $1.5\times$), without any utility drop, if we only compute the output gradient but not the parameter gradient during the backward propagation.
\end{abstract}

\section{Introduction}

Adversarial robustness is an important field to protect models from adversarial attacks, since deep neural networks that are highly accurate can be vulnerable to small input perturbations. For example, the projected gradient descent (PGD) attack can worsen the ResNet50, from 95.25\% accuracy to 0.00\% using a small perturbation of $l_\infty$ magnitude 8/255 on CIFAR10, as well as from 76.13\% accuracy to around 0.01\% under the same attack on ImageNet \cite{madry2018towards}. This vulnerability is observed not only in computer vision but also in natural language processing, e.g. a word-level attack TextFooler \cite{jin2020bert} has 90\% success rate of fooling BERT \cite{kenton2019bert} on SST-2 dataset by \cite{zeng2021openattack}. At the core of adversarial robustness, one need the adversarial perturbation to construct adversarial examples, which are used to evaluate a model's robustness and to defend against attacks via the adversarial training (sometimes improving the accuracy as well).

Mathematically speaking, the adversarial perturbation is the solution to a constrained maximization problem in the input space:
\begin{align}
\p^*=\text{arg max}_{\p\in \mathcal{B}} L(f(\x+\p;\W),\y)
\label{eq:maximization adv}
\end{align}
where $\x$ is the data feature, $\y$ is the label, $\W$ is the model parameters, $f$ is the model function, $L$ is the loss function, $\p^*$ is the adversarial perturbation in the attack domain $\mathcal{B}$, e.g. $\ell_\infty$ ball with some radius $\epsilon$ known as the attack magnitude. Many optimizers (known as the attack methods) including the fast gradient sign method (FGSM; \cite{goodfellow2014explaining}) and PGD have been proposed to iteratively compute $\frac{\partial L}{\partial \p}$ and to solve \eqref{eq:maximization adv}. These methods have achieved empirical success over the decades.
\begin{algorithm}[!htb]
   \caption{Adversarial perturbation by optimization}
   \label{alg:adv}
\begin{algorithmic}
   \STATE {\bfseries Input:} data $\x$, label $\y$, model parameter $\W$
   \FOR{$k=1,2,\cdots,K$}
   \STATE forward propagation $\hat\y=f(\x+\p;\W)$
   \STATE backward propagation $\nabla_{\p} L=\frac{\partial L(\hat\y,\y)}{\partial \p}$
   \STATE input update $\p=\text{projection}_{\mathcal{B}}(\p+\eta_k\nabla_{\p} L)$ 
   \ENDFOR
\end{algorithmic}
\end{algorithm}

Generally speaking, the optimization \eqref{eq:maximization adv} is non-convex and hard-to-solve. Therefore, strong adversarial perturbation is usually solved by applying the attack methods for a number of steps, i.e. $K\geq 1$. However, this approach may cause a heavy computational burden, e.g. about $20\times$ time complexity in the 20-step PGD adversarial training, or $2\times$ time complexity in FGSM adversarial training, in comparison with the standard non-robust training.

\subsection{Contributions}
We propose a trick, the semi-backward, to reduce the time complexity of backward propagation to half. We analyze the effectiveness of the semi-backward trick in theory and in practice across different attacks and libraries, all of which only require to add a single line of code.

\subsection{Organization}
In Section 2, we will discuss the backward propagation and its 2 processes, with a brief introduction of their time complexity. In Section 3, we propose to only compute 1 out of 2 processes: we can do semi-backward propagation by only computing the output gradient. This is valid by the chain rules and implemented by adding \texttt{[p.requires\_grad\_(False) for p in model.parameters()]}. In Section 4, we experiment with state-of-the-art adversarial robustness libraries and attack methods to demonstrate the advantage of semi-backward propagation. Notice that no accuracy results are provided, because we focus on the efficiency and the accuracy is the same for both full backward and semi-backward propagation. At last, we discuss some details about the Pytorch implementation of adversarial perturbation.  

\section{Preliminaries on a Single Layer} 
\subsection{Forward and backward propagation}
To understand and then improve the computation of adversarial perturbation, we start from the standard training. The forward propagation on the $l$-th layer\footnote{Here the demonstration is on a linear layer for simplicity, which can easily extend to other layers.} is:
$$\a_{l+1}=\phi_{l+1}(\s_{l}) \text{ and } \s_{l+1}=\a_{l+1}\W_{l+1}+\b_{l+1}$$
Here $\a$ is the layer's input (i.e. the activations), $\s$ is the layer's output (i.e. the pre-activation), $\W,\b$ are the layer's weights and biases, $\phi$ is the inter-layer operation such as the activation functions (ReLU, tanh, etc.) and the pooling. 

The backward propagation of the $l$-th layer has 2 underlying processes: 
\begin{enumerate}
    \item the computation of \textbf{output gradient}
\begin{align}
\frac{\partial {L}}{\partial \s_{l}}=\frac{\partial {L}}{\partial \s_{l+1}}\W_{l+1}\circ \phi'(\s_{l}),
\label{eq:output grad}
\end{align}

\item the computation of \textbf{parameter gradient}
\begin{align}
\begin{split}
\frac{\partial {L}}{\partial \W_{l}}&=\frac{\partial {L}}{\partial \s_{l}}^\top\frac{\partial \s_{l}}{\partial \W_{l}}=\frac{\partial {L}}{\partial \s_{l}}^\top\a_{l},
\\
\frac{\partial {L}}{\partial \b_{l}}&=\frac{\partial {L}}{\partial \s_{l}}^\top\frac{\partial \s_{l}}{\partial \b_{l}}=\frac{\partial {L}}{\partial \s_{l}}^\top\mathbf{1},
\end{split}
\label{eq:param grad}
\end{align}
\end{enumerate}

\begin{figure}[!htb]
    \centering
    \includegraphics[width=0.3\linewidth]{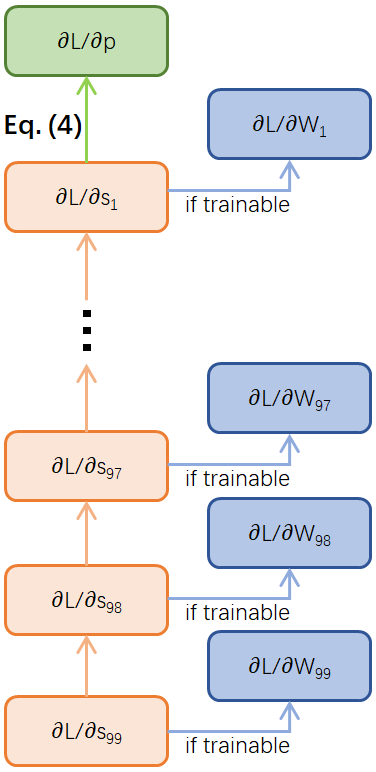}
    \caption{Computation graph of backward propagation.}
    \label{fig:my_label}
\end{figure}

The derivation for both computations is obvious from the chain rules. We will visualize these computations in sequential order in \Cref{fig:my_label}.

\subsection{Complexity of propagation}

It is well-known that the forward propagation and each process in the backward propagation have the same time complexity (up to lower order terms\footnote{It suffices to state the highest order term here as it dominates the time complexity, e.g. if the layer maps from $\R^{BTd}$ to $\R^{BTp}$, the computation of output gradient is $2BTdp+2BTd\approx 2BTdp$.}), as summarized in \Cref{fact:two process same time}.

\begin{fact}\label{fact:two process same time}
For one layer, the forward propagation (i.e. the computation of one layer's input $\a$ and output $\s$), the computation of output gradient and the computation of parameter gradient have (almost) the same time complexity as $2BTM$, where $B$ is batch size, $T$ is hidden feature dimension, and $M$ is the number of parameters in this layer.
\end{fact}
\begin{remark}
Here $T$ is the sequence length for text data (e.g. sentence length); $T$ is the dimension product (height $\times$ width) of hidden feature for image data.
\end{remark}

\section{Semi-backward Propagation for Adversarial Perturbation}
By \Cref{fact:two process same time}, the optimization of adversarial perturbation (forward and backward) has a time complexity $6BTM$. This can be reduced to $4BTM$ but still give the same adversarial perturbation:
$$\underbrace{\text{forward propagation }}_\text{time complexity $2BTM$} + \underbrace{\text{ output gradient }}_{2BTM} + \underbrace{\cancel{\text{ parameter gradient}}}_{2BTM}$$
We term this pipeline as the semi-backward propagation.

\subsection{A computation graph viewpoint}
We claim that during the optimization in \Cref{alg:adv}, only the output gradient is needed, thus the time complexity of backward propagation can be halved since we eliminate 1 out of 2 processes. To see this, we present the computation graph of all layers and leverage the single-layer results from the previous section.

Note that the output gradient is computed via \eqref{eq:output grad} (orange arrow), the parameter gradient is computed via \eqref{eq:param grad} (blue arrow), and perturbation gradient is computed via \eqref{eq:data grad} (green arrow).

\begin{align}
\begin{split}
\textbf{Forward: }&
\s_1=(\x+\p)\W_1+\b_1
\\
\textbf{Backward: }&
\frac{\partial {L}}{\partial \p}=\frac{\partial {L}}{\partial \s_{1}}^\top\frac{\partial\s_{1}}{\partial \p}=\frac{\partial {L}}{\partial \s_{1}}^\top\W_1.
\end{split}
\label{eq:data grad}
\end{align}

\subsection{Reduced Complexity and Speedup}
\begin{fact}
For a neural network, the time complexity of computing the adversarial perturbation gradient is $6B\sum_l T_l M_l$ without semi-backward propagation, and $4B\sum_l T_l M_l$ with semi-backward propagation.
\end{fact}

In terms of training speed, the improvement is $4/2=2\times$ on the backward propagation and $6/4=1.5\times$ overall. This complexity reduction is for each step of perturbation optimization, and thus the speedup holds for multi-step and single-step attacks.

Although the improvement is mainly on the time complexity, the semi-backward has some additional improvement on the space complexity, e.g. about 5\% reduction of memory cost for the PGD experiment in \Cref{sec:PGD benchmark} using Torchattacks \cite{kim2020torchattacks}.

\subsection{Code snippet}

This high-level idea is applicable to the general auto-differentiation frameworks, such as Pytorch \cite{paszke2019pytorch}. We provide a code snippet based on \texttt{torchattacks}(v3.3.0)\cite{kim2020torchattacks}, \texttt{timm}(v0.6.11)\cite{rw2019timm} and \texttt{torch}(v1.12)\cite{paszke2019pytorch}, that adds only one line (in red text) to implement the semi-backward propagation.

\begin{Verbatim}[commandchars=\\\{\}]
import timm, torchattacks
model=timm.create_model('resnet50')
image=torch.rand(size=(16,3,224,224))
label=torch.Tensor([1]*16).long()
\textcolor{red}{[p.requires_grad_(False) for p in model.parameters()]}
atk = torchattacks.PGD(model)
adv_images = atk(image,label)
\end{Verbatim}

\section{Experiments}
We evaluate the effectiveness of semi-backward propagation, which enjoys a theoretical speedup by $1.5\times$, on widely-used Python libraries in their latest version: Torchattacks (v3.3.0)\cite{kim2020torchattacks}, Adversarial-Robustness-Toolbox (ART; v1.12.1)\cite{art2018}, AdverTorch (v0.2.4)\cite{ding2019advertorch}, DeepRobust (v0.2.5)\cite{li2020deeprobust}, Foolbox (v3.3.3)\cite{rauber2017foolboxnative}, CleverHans (v4.0.0)\cite{papernot2018cleverhans}, and OpenAttack (v2.1.1)\cite{zeng2021openattack}. We use 1 Tesla T4 GPU and record the time in seconds.

\subsection{PGD benchmarks}
\label{sec:PGD benchmark}

We use 16 ImageNet images of $3\times 224\times 224$ with different libraries. We attack the ResNet50 model (62 million parameters, \cite{he2016deep}) with PGD. We do not report the choice of attack domain $\mathcal{B}$, attack magnitude nor learning rate, because these hyperparameters do not affect the efficiency and the accuracy will be the same with/without semi-backward propagation.
\begin{table*}[!htb]
    \centering
    \resizebox{0.9\linewidth}{!}{
    \begin{tabular}{c|c|c|c|c|c|c}
         &Torchattacks& Foolbox &ART&AdverTorch& DeepRobust &CleverHans
         \\\hline
         original&6.89$\pm$0.11&7.38$\pm$0.08&7.09$\pm$0.03&7.43$\pm$0.08&7.23$\pm$0.07&7.36$\pm$0.09
         \\
         semi-backward&4.74$\pm$0.02 &5.44$\pm$0.02&5.00$\pm$0.06&5.48$\pm$0.02&4.94$\pm$0.01&4.96$\pm$0.03
         \\
         speedup&1.45$\times$ &1.36$\times$&1.44$\times$&1.36$\times$&1.46$\times$&1.48$\times$
    \end{tabular}
    }
    \caption{Time over 10 runs ($\pm$ standard deviation) for adversarial perturbation of 50-step PGD on ResNet50 with 16 images. The speedup remains constant for different number of PGD steps.}
    \label{tab:pgd}
\end{table*}

We also test different batch sizes and model architectures, including Vision Transformer (ViT \cite{dosovitskiy2020image}, 86 million parameters) on Torchattacks \cite{kim2020torchattacks}.
\begin{table*}[!htb]
    \centering
    \resizebox{0.8\linewidth}{!}{
    \begin{tabular}{c|c|c|c|c|c|c}
          Batch size (ResNet50)&4&8&16&32&64&128
          \\\hline
          original&2.06&3.99&6.89&13.83&27.24&54.6
         \\
         semi-backward&1.41&2.75&4.74&9.44&18.60&36.6
         \\
         speedup&1.45$\times$ &1.45$\times$&1.45$\times$&1.46$\times$&1.46$\times$&1.49$\times$
         \\\hline\hline
          Model (Batch size 16)&ResNet18&ResNet34&ResNet50&ResNet101&ResNet152&ViT-base
          \\\hline
          original&2.13&3.57&6.89&11.92&16.49&26.5
         \\
         semi-backward&1.52&2.54&4.74&7.89&11.01&19.2
         \\
         speedup&1.40$\times$&1.41$\times$&1.45$\times$ &1.50$\times$&1.50$\times$&1.38$\times$         
    \end{tabular}
    }
    \caption{Average time over 10 runs for adversarial perturbation of 50-step PGD on ResNets with different batch sizes, using Torchattacks.}
    \label{tab:pgd2}
\end{table*}

\subsection{Attack on images}
We test different attacks on ResNet50 using ART and Torchattacks (shorthanded as TA), covering AutoAttack \cite{croce2020reliable}, BIM \cite{kurakin2018adversarial}, Sqaure \cite{andriushchenko2020square}, APGD \cite{croce2020reliable}, CW \cite{carlini2017towards}, MIFGSM \cite{dong2018boosting}, SINIFGSM \cite{lin2019nesterov}, and VMIFGSM \cite{wang2021enhancing}. For each attack we use the default hyperparameters in the corresponding library.
\begin{table*}[!htb]
    \centering
    \resizebox{\linewidth}{!}{
    \begin{tabular}{c|c|c|c|c|c||c|c|c|c|c||c}
         &PGD&AutoAttack&BIM&Square&FGSM&APGD&CW&MIFGSM&SINIFGSM&VMIFGSM&\multirow{2}{*}{Average}
         \\\cline{1-11}
         library&ART&ART&ART&ART&ART& TA& TA& TA& TA& TA
         \\\hline
         original&15.2&0.08&15.2&0.14&0.20&0.07&6.85&1.18&6.81&8.27&-
         \\
         semi-backward &10.2&0.06&10.2&0.11&0.15&0.05&4.76&0.94&4.71&5.68&-
         \\
         speedup&1.48$\times$&1.25$\times$&1.49$\times$&1.27$\times$&1.31$\times$&1.35$\times$&1.44$\times$&1.26$\times$&1.45$\times$&1.46$\times$&1.38$\times$
    \end{tabular}
    }
    \caption{Average time over 10 runs for adversarial perturbation of attacks with default setting on ResNet50 with 16 images.}
    \label{tab:nlp}
\end{table*}

\subsection{Attack on texts}
We test different attacks on BERT (109 million parameters) using the OpenAttack library\footnote{See \url{https://github.com/thunlp/OpenAttack\#attack-built-in-victim-models}.}, covering TextFooler \cite{jin2020bert}, PWWS \cite{ren2019generating}, Genetic \cite{alzantot2018generating}, SememePSO \cite{zang2020word}, BERTAttack \cite{li2020bert}, FD \cite{papernot2016crafting}, DeepWordBug \cite{gao2018black}, and VIPER \cite{eger2019text}. For each attack we use the default hyperparameters in the corresponding library.
\begin{table*}[!htb]
    \centering
    \resizebox{\linewidth}{!}{
    \begin{tabular}{c|c|c|c|c|c|c|c|c||c}
         &\multicolumn{6}{|c|}{Word-level}&\multicolumn{2}{|c|}{Character-level}&\multirow{2}{*}{Average}
         \\\cline{1-9}         
         &TextFooler& PWWS& Genetic&SememePSO&BERTAttack&FD& DeepWordBug &VIPER
         \\\hline
         original&90.4&113.9&918.7&387.8&147.2&198.1&70.7&237.1&-
         \\
         semi-backward &65.9&80.1&603.3&312.9&114.4&158.0&53.5&184.5&-
         \\
         speedup&1.37$\times$&1.42$\times$&1.52$\times$&1.24$\times$&1.29$\times$&1.25$\times$&1.32$\times$&1.29$\times$&1.34$\times$
    \end{tabular}
    }
    \caption{Average time over 10 runs for adversarial perturbation of attacks with default setting on BERT with 20 examples.}
    \label{tab:nlp}
\end{table*}

\subsection{Adversarial training}
We extend our discussion to the adversarial training with $K$-step attacks:
\begin{align}
\min_\W \max_{\p\in \mathcal{B}} L(f(\x+\p;\W),\y)
\label{eq:adv train}
\end{align}

\begin{figure}[!htb]
    \centering
    \includegraphics[height=7cm]{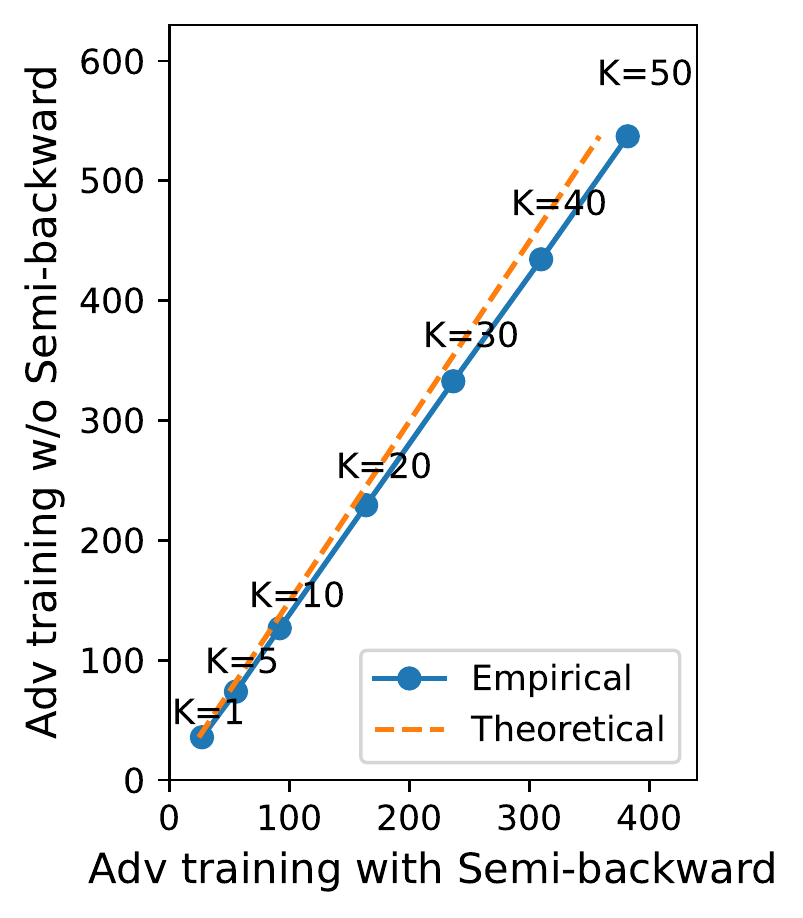}
    \caption{Per-epoch training time with/without semi-backward propagation in adversarial training, using ResNet18 on CIFAR10.}
    \label{fig:train}
\end{figure}

We claim that, without semi-backward propagation, the time complexity of one iteration (1 $\W$ update, $K \p$ updates) is $(6*1+6*K)B\sum_l T_l M_l$; with semi-backward propagation, the time complexity of one iteration is $(6*1+4*K)B\sum_l T_l M_l$. Therefore, the theoretical speedup is $\frac{6K+6}{4K+6}$. We demonstrate the empirical speedup in \Cref{fig:train}, using the example in DeepRobust \url{https://github.com/DSE-MSU/DeepRobust\#image-attack-and-defense} with CIFAR10 dataset.
To use semi-backward propagation in adversarial training but full backward propagation during the parameter update, we need to turn on and off the computation of parameter gradient, as demonstrated in Appendix.

\section{Discussion}
In this paper, we show that using semi-backward propagation in place of full backward propagation can significantly accelerate the adversarial perturbation, as well as slightly reduce the memory cost. Notice that in Pytroch, turning off \texttt{param.requires\_grad} is different from using \texttt{torch.autograd.grad(loss,$\p$)} or \texttt{loss.backward(inputs=$\p$)}\footnote{These approaches have been used in DeepRobust, ART, Torchattacks, etc., which are still accelerated by semi-backward trick in \Cref{tab:pgd}.}, where $\p$ is the adversarial perturbation: the first approach applies before the forward propagation to define the computation graph.

\bibliography{example_paper}
\bibliographystyle{plain}

\appendix
\section{Pseudo-code for adversarial training with semi-backward trick}
\begin{Verbatim}[commandchars=\\\{\}]
import timm, torchattacks
model=timm.create_model('resnet50')
image=torch.rand(size=(16,3,224,224))
label=torch.Tensor([1]*16).long()
for t in range(999):
    \textcolor{red}{# turn off parameter gradient}
    for p in model.parameters():
        p.initial_requires_grad=p.requires_grad
        p.requires_grad_(False)
    atk = torchattacks.PGD(model, steps=K)
    image = atk(image,label)
    
    \textcolor{red}{# turn on parameter gradient}
    for p in model.parameters():
        p.requires_grad=p.initial_requires_grad
    loss = F.cross_entropy(model(image),label)
    loss.backward()
    optimizer.step()
    optimizer.zero_grad()
\end{Verbatim}
\end{document}